\documentclass[runningheads]{llncs}
\usepackage{amsmath}
\usepackage{wrapfig}
\usepackage[paperheight=11in,
   paperwidth=8.5in,
   top=25mm,
   bottom=25mm,
   left=25mm,
   right=25mm]{geometry}
\usepackage{longtable}
\usepackage[T1]{fontenc}
\usepackage{float}
\usepackage{graphicx}
\usepackage{array} 
\usepackage{xcolor}
\usepackage[linktocpage, colorlinks, linkcolor=blue, citecolor=blue]{hyperref}
\usepackage{natbib}

\usepackage{rotating}

\usepackage{orcidlink}
\usepackage{amssymb}

\begin{document}

\title{Safe Continual Reinforcement Learning Methods for Nonstationary Environments. Towards a Survey of the State of the Art
\thanks{if necessary}
}

\author{Timofey Tomashevskiy}
\authorrunning{T. Tomashevskiy et al.}

\institute{McMaster Centre for Software Certification, Department of Computing and Software, McMaster University, Canada\\  \email{ tomashet@mcmaster.ca }
}

\maketitle          
\begin{abstract}

Safety is a vital component for the future development and acceptance of reinforcement
learning (RL) methods. Although over the last decades, online reinforcement learning and learning in nonstationary environments have made huge progress, safe continual online reinforcement learning remains one of the most challenging topics of research. Despite numerous recent works in the safe RL field and in continual online RL learning, a systematic understanding of how to establish safety during continual online learning under nonstationary conditions remains limited. This happened because of the complexity of 
problem that requires a combination of knowledge from different research areas, including safe learning, adaptation to the distribution shift, and safe optimization. Combination of knowledge from these domains is not a straightforward task because of numerous challenges 
including the consecutive nature of data flow, potential delay of the reward, unpredictable nature of the nonstationarity (NS) dynamics, and difficulties in precise constraint formulation.  In this article, we provide a review of existing 
continual online safe reinforcement learning (COSRL) methods. 
We provide the taxonomy of COSRL based on the type of safe learning mechanism that takes adaptation to nonstationarity into account. We categorise safety constraints formulation for online reinforcement learning algorithms, and finally, we discuss challenges and prospects for creating reliable, safe online learning methods.   

\end{abstract}

\section{Introduction}

\label{sec:introduction}
Reinforcement learning is a method for solving sequential tasks when the agent perceives the states and acts to maximize the long-term return, which is based on a real-valued reward \cite{altman1999constrained}. In real life, aside from the return, safety comes into play. The potential users of the algorithms should be confident not just in performance but also in the safety of the method they apply. While there are many definitions of safety, most of them take uncertainty and risk consideration into account \cite{garcia2015comprehensive}. 
Therefore, to avoid or limit the risk and guarantee safety for traditional RL methods, the paradigm of RL was extended, and different types of safety constraints were introduced. Safe reinforcement learning has made great progress during the last decades and is now a relatively mature discipline able to provide 
safety constraints satisfaction guarantees in many settings.  
However, despite the progress achieved, acceptance of RL applications in the real world remains limited. One of the reasons for that is the nonstationary stochastic nature of the real world. Different forms of non-stationarity can  
affect the performance and compromise the safety of the learning agent. 
Distribution shift is one of the most common forms of non-stationarity faced by RL. It appears when training and testing distributions are different. Most existing safe RL (SRL) methods are based on the assumption of stationarity of the environment and underlying Markov process, where the state, reward, action space distribution, and transition dynamics of the environment remain stationary and don't change over time. In practice, all these parameters can show some time dependence. Nonstationarity makes it difficult to directly apply traditional stationary reinforcement learning methods in potentially nonstationary environments without some adjustment to nonstationarity conditions. While existing safe online reinforcement methods based on the assumption of stationarity\cite{berkenkamp2015safe},\cite{daulton2022robust} can guarantee some robustness to non-stationarity, it is not always clear to what extent initial safety constraints can be respected during and after the distribution shift. Successful adjustment of safe reinforcement learning to nonstationary environments requires adaptation of both: the performance of the agent, i.e, adaptation of the reinforcement learning mechanism, and adaptation of the safety constraints,  as initial constraints can become inefficient or unsafe in a new, changed environment. \\
Adaptation of unconstrained RL methods to nonstationarity has been a subject of intensive research over the last few years, including continual learning \cite{brunskill2014pac,chandak2020lifelong,parisi2019continual,abel2018policy} and meta-learning \cite{al2017continuous,xie2020deep,vanschoren2019meta} directions.  
At the same time, the number of studies specifically focused on safe continual reinforcement learning under non-stationarity remains limited because of the complicated nature of the subject that lies at the intersection of sequential learning, adaptation to non-stationarity, and safety constraints satisfaction.

The research studies related to adaptation to nonstationarity, while satisfying constraints, are scattered across different domains of RL, including RL meta-learning \cite{wang2023enforcing,finn2017model,rakelly2019efficient}, methods of predictive control, constrained Markov decision processes (CMDP) \cite{altman1999constrained,altman2021constrained,borkar2014risk,wachi2020safe}, and Bayesian exploration/optimization \cite{berkenkamp2023bayesian, ghavamzadeh2015bayesian,mockus2002bayesian,mockus2005bayesian}. Existing works study adaptation to different forms of NS, such as active, passive, and mixed NS \cite{chandak2022reinforcement}. Most of them are primarily focused on the adaptation and high performance of the learning agent, rather than on satisfying safety constraints. Some of them respect constraint satisfaction but don't pay much attention to performance or to speed of NS-adaption. We aim to bring knowledge of learning safety and adaptation together to shed light on COSRL that can facilitate future research.

\subsection{Our contribution}

In this paper, we review and evaluate current state-of-the-art approaches to the safety of continual online RL under non-stationarity. 
We discuss existing adaptation mechanisms, optimization methods, and types of safety constraints. We provide a taxonomy of existing safe online learning methods based on the adaptation mechanism. We also provide a taxonomy and detailed review of the state-of-the-art safety constraints formulation and discuss possible directions of constraints formulation for continual online safe RL methods. 
\subsection{Related Studies}
\label{sec: Related Studies}
Safe reinforcement learning is a general and flexible concept that covers optimization and safety and exists in many different settings, including online and offline learning, model-based, model-free, stationary, or nonstationary, to name just a few. Several previous papers covered different aspects of safe RL. The comprehensive survey of Garcia and Fernandez \cite{garcia2015comprehensive} covers safety approaches and provides a taxonomy of safe RL algorithms. Shangding et al. \cite{gu2022review} covered five fundamental problems of safe RL and problems related to the practical application of reinforcement learning.  Embedding soft constraints in policy optimization was covered by \cite{altman2021constrained,wachi2020safe, kim2020safe} surveys. \cite{liu2021survey} reviews the state of research in constrained policy optimization for model-free algorithms.   Several research works covered particular methods in RL, classified by the type of learning or by the type of applications. \cite{hospedales2021meta} reviewed safe meta-learning methods. \cite{brunke2022safe} provided a survey of safe RL methods in Robotics.  All these studies focus on safety for reinforcement learning methods based on the stationarity assumption of the environment, and don't cover safe continual online learning under nonstationarity. \cite{padakandla2021survey, khetarpal2022towards} outline approaches for RL adaptation for non-stationarity, but don't tackle safety constraints during or after the adaptation. Concise but very well focused on safety constraints \cite{wachi2024survey} provides a great review of constraints formulation; however,  this work does not specifically cover constraints formulation for nonstationary environments or online learning algorithms.  Our survey aims to fulfill these gaps and provide a review of the state-of-the-art methods and theoretical aspects of building safe continual reinforcement learning algorithms able to work in nonstationary environments. 
\subsection{Overview}
\label{sec: Overview}

The rest of this review is organized as follows.  In Section \ref{sec: Preliminaries}, we provide preliminaries for safe continual reinforcement learning. We provide definitions, notations, and explain the main concepts required for understanding COSR. 
Section 3 provides problem formulation. In this section, we explain three key components of COSRL.
Section 4 describes challenges of building safe online RL algorithms.Section 5 categorises current solutions based on safety adaptation techniques.
Section 6 provides the details  
of the most important COSRL works. 
Section 7 provides a taxonomy of constraint formulation for COSRL. Section 8 concludes the study by discussing findings and open problems.

\subsection{Scope of Review}
In this review, we present a survey of research on the safety of online reinforcement learning under non-stationarity.  We cover the optimization, adaptation, and safety aspects of the algorithms, but we don't address the catastrophic forgetting problem. We focus on a single learning agent setting and don't consider multi-agent learning, as we believe that both topics require separate research.   

\section{Preliminaries}
\label{sec: Preliminaries}
In this section, we provide notations, definitions, and main ideas that can help better understand the general concept of COSRL.
\subsection{Notations}
In this survey, capital letters are used for random variables, while lower-case letters are used for the values of random variables and for scalar functions. For consistency with prior literature, we largely follow the notation of (Sutton and Barto, 1998)\cite{sutton1998reinforcement}.

\subsection{Markov Decision Process (MDP)}
 Most of the existing reinforcement learning methods are based on the Markov decision process (MDP) that can be represented as a tuple 
$(S, A, R, P, \Psi)$ where S is the set of states, A is the
set of actions, R : S × A × S → R is the reward function,
P: S×A×S → [0, 1] is the transition probability function
where P($s_{i+1}$| $s_{i}$, a) is the probability of transitioning to state
$s_{i+1}$ given that the previous state was $s_{i}$ and the agent took
action a in $s_{i}$; and $\Psi$ : S → [0, 1] is the starting state
distribution \cite{bellman1957markovian}. A stationary policy $ \pi \in\Pi $  maps
from states to probability distributions over actions $ \pi $ : S → P(A), with $ \pi$(a|s) denoting the probability of selecting action a in
state s. $ \Pi $ denotes the set of all stationary policies. 
\cite{sutton2018reinforcement}. Each learning task can be represented as some  
\begin{equation}
   MDP_i(S, A, R, P, \Psi  ) \label{eq0} 
\end{equation} \label{eq0}
where $i$ represents episode number, and the goal of the learning agent is to find such a policy $\pi$,   that maximizes the objective function $J_{\pi}$ over the episodes $ i-N$. 

\begin{equation} \label{eq:eq1}
\max_{\pi\in \Pi} J_{\pi} =  \max_{\pi\in \Pi} \hspace{0.2 cm}  \mathbb {E}_{r_\pi}\Big[  \sum_{t=0}^{\infty} \gamma^t *r_{t} \Big] 
\end{equation} where $r_{t}$ represents the reward,  and $\gamma^t$ represents the discount factor.
However, such a clear and convenient formulation does not take any goals other than performance into account. In real life, the agent can have multiple goals and multiple objectives that should be satisfied at the same time. One of the most important goals that is different from performance is safety.\\
Although the definition of RL safety varies from one study to another, in this work, we define safety as the state of being protected from harm or other dangers.\\
We also define safe learning as the process of learning policies that maximize the expected return 
while satisfying safety constraints during the learning and deployment processes \cite{garcia2015comprehensive}. 
So, we can modify equation (\ref{eq:eq1}) by adding some constraints C on policy $\pi$ aimed to represent the safety part.
\begin{equation}\label{eq:eq2}
 \max_{\pi\in \Pi} J_{\pi} =  \max_{\pi\in \Pi} \hspace{0.2 cm}  \mathbb {E}_{r_\pi}\Big[  \sum_{t=0}^{\infty} \gamma^t *r_{t} \Big]
 \hspace{0.5 cm}
 s.t.\hspace{0.2 cm} C{_\pi}= \mathbb {E} _{c _ \pi}\Big[ \sum_{t=0}^{\infty} \gamma ^ t*c_{t}\Big] \le d   \ \ \ \space      
\end{equation}
Where $C_{\pi}$ represents the safety constraint function that shows the level of expected cumulative cost value 
and $d$ represents the safety threshold. In the future, we would call $C_\pi$ a cost function. Now, we can represent a modified version of MDP  - constraint Markov decision process, or CMDP \cite{altman2021constrained} that takes constraints into account  and can be formulated as a tuple: 
\begin{equation} \label{eq:eq4}
   CMDP_i(S ,A ,R ,P ,\Psi , C )  
\end{equation}
\subsection{Nonstationarity}
Reinforcement learning algorithms don't work well on unfamiliar data distributions. If the training and testing data are different, the performance of the algorithm would be low. 
Both MDP and CMDP are based on the stationarity assumption on the underlying distributions of $S, P, A, R, C$. However, this assumption does not hold for most practical applications where underlying processes are nonstationary. To make RL algorithms more applicable, the influence of nonstationarity should be taken into account.
We define nonstationarity as a change of distribution of at least one of the MDP parameters from one task to another.
\textbf{Types of Nonstationarity (NS)} 
Nonstationarity can be classified in several ways. One way to classify NS is by its origin.  

\textbf{Origin of NS} can be split into three groups:  passive, or exogenous, non-stationarity caused by external factors, active non-stationarity caused by the learning agent itself, and a hybrid type which is a combination of the two previous types\cite{chandak2022reinforcement}. 

\textbf{Type of Distribution Shift (DS)}
Nonstationarity can also be classified by the type of distribution shift. Distribution shift occurs when the joint distribution of inputs $ x \in X$ and outputs $ y \in Y$  differs between training and test stages, i.e.
\begin{center}
    $P_{train}( x, y ) {\neq} P_{test} ( x,y )$ \\
\end{center}
There are four main types of distribution shift: covariate shift, label shift, concept drift, and domain shift. 
\textbf{Covariate shift} happens when the distribution of the input data, $p(x)$, changes, but the conditional distribution of the output given the input, $P(y| x)$, remains the same.
\textbf{Label shift}, or so-called prior probability shift, occurs when there is a significant change in the distribution of the target variable $P(y)$. Mathematically, the label shift happened when the class label distribution $P(y)$ changes, but the class-conditional distribution $p(y|x)$ remains unchanged. 
\textbf{Concept drift} refers to the change in the conditional distribution $ P(y|x)$, such as relations between the input features $X$ and the target variable $Y$. Finally, in \textbf{domain shift}, or so-called \textbf{joint distribution shift}, the distribution of inputs, $P(x)$, and the conditional distribution of outputs given inputs, $P(y|x)$, both change. 
In reinforced learning, DS can be represented by utilizing action $ a \in A $ of the learning agent as an input variable and reward $r\in R$ as the target variable. Alternatively, we can use states $s\in S$ as input and transition probability $p\in P$ as the target variables.
 Most typical changes considered by the research are the changes in the distribution of the reward $R$, changes in the transition probability function $P$, and changes in the distribution of the action space  $A$. Even though in the general case all MDP elements can have time dependence, we are not aware of the work that studies these settings.
\subsection{Nonstationary MDPs}
A number of studies that investigate adaptation to non-stationarity try to modify the Markov decision process model so that it can better reflect non-stationarity. These works can be divided into two groups: POMDP-based and MDP-based methods.
\subsubsection{POMDP}
The first group uses the partially observable Markov decision process (POMDP), the framework proposed by Åström \cite{aastrom1965optimal} and later developed by Kaelbling \cite{kaelbling1998planning}. This is a broad framework that allows the representation of uncertainty and non-stationarity of the environment, when non-stationarity can be seen as the agent's understanding of the environment rather than the property of the environment. The broad nature of the POMDP makes it attractive for modeling non-stationarity on a high level, but brings some difficulties in implementation. There were several efforts to extend POMDP to simplify its implementation
for nonstationary environments.
\subsubsection{HM-MDP}
Choi et. al \cite{choi2001hidden} developed the Hidden Mode Markov Decision Process \textbf{(HM-MDP)}, a special case of the POMDP model, that represents the non-stationary environment as a set of stationary environments, also known as a set of \textit{contexts}, or modes.  
HM-MDP assumes that the number of hidden modes, or the contexts, is known, and proposes a way of learning the HM-MDP model by using the Baum-Welch algorithm. Da Silva et. al \cite{da2006improving} propose methods that can detect a prior unknown set of contexts. The Bayes-adaptive {(BAMDP)} model was developed by \cite{ross2007bayes},\cite{duff2002optimal}. Both HM-DP and BAMDP consider an episodic MDP with an unknown reward and transition dynamics that have to be found during the episode. \cite{schope2021constrained} proposed a constrained POMDP formulation and algorithms. \\
The second group includes research that extends traditional stationary MDPs \cite{puterman1990markov} to non-stationary environments.\\
\subsubsection{NSMDP} 
Lecarpentier et. al \cite{lecarpentier2019non} proposed 
NSMDP - a modified version of MDP that represents nonstationary MDP $M_{NS}$  as a time series of stationary MDPs $M_{S_1}, M_{S_2} ... M_{S_k}$ where $ k$ is an epoch number. The distribution of transition function P and reward functions R can slowly vary over time between MDPs. Method assumes Lipschitz continuity of both transition and reward functions. Assumption of slow change allows to plan efficiently within NSMDPs. Based on the proposed approach authors designed a worst-case approach algorithm that allows to select the best of the possible worst-case scenarios.
\cite{lecarpentier2019non} allows us to reason about NS dynamics, and on the other hand, it allows us to use all set of RL methods developed for stationary MDPs.
Even though \cite{aastrom1965optimal,choi2001hidden, lecarpentier2019non} frameworks have proven their effectiveness, they do not actively predict the dynamic of the 
environment between the episodes. That makes the adaptation process reactive rather than proactive and can affect the performance of the agent.
\subsubsection{DP-MDP}
Xie et. al \cite{xie2020deep} built a Dynamic Parameter model \textbf{(DP-MDP)} that corrects this shortcoming. The proposed model allows reinforcement learning agents not only to learn the latent variables but also to predict their dynamics and proactively adjust to the upcoming changes.

\section{Problem Formulation}

In this section, we introduce problem formulation for continual online safe reinforcement learning (COSRL) under nonstationarity and introduce the notations that we will use in this paper. 
Continual online reinforcement learning (COSRL) in nonstationary environments research is at the intersection of three research areas: online learning, adaptation to nonstationarity, and safe reinforcement learning.
We are going to consider the existing research based on these three dimensions.
\subsection{Learning in a nonstationary environment}
In this work, we consider episodic MDP settings where each task represents a separate episode, an MDP process $M_i$, $i\in [1,n]$ where $i$ denotes an episode index, and $n\in N$ is the number of episodes. Change of episodes occurs at some changepoints $T_{ch}$. The length, the number, and the sequence of episodes are unknown a priori. So, we have a series of episodes $M_1, M_2 ..... M_n$, where each episode can be represented as a stationary MDP $M_i(S_i, A_i, R_i, P_i, \Psi_i) $. MDP parameters can change between episodes. Nonstationarity can be seen as the difference between distributions of separate parameters of different MDPs, e.g. transition probability function or the reward. In a more practical setting, nonstationarity can be seen as the difference between tasks \cite{visus2021taxonomy},\cite{khattar2024cmdp},\cite{song2016measuring}. The goal of the learning agent is to find such a policy $\pi$ that would maximize the expected cumulative reward collected while travelling over tasks from 1 to $n$.  
\begin{equation} 
\label{eq:eq5}
  \max_{\pi\in \Pi} J_{\pi} =  \max_{\pi\in \Pi}\mathbb{E}_{\pi} \left (\sum_{i=1}^n \sum_{t=1}^\tau \gamma^t *r_{t} \right ) 
\end{equation} 
where $J_{\pi}$ is an objective function, $\pi \in \Pi$ is a policy, $i$ is an episode index, $n\in N$ is the number of episodes, and $\tau$ is the number of time steps.\\
\subsection{Adaptation to nonstationarity} 
As a learning agent operates in nonstationary conditions, it is important to take the effect of nonstationarity into account. Speed of adjustment to NS is an important factor for both performance and safety. For example, if the adjustment to an unfamiliar environment takes too much time for the control system of an autonomous vehicle, the vehicle can crash.\\
Let's consider how quickly the agent can adjust to some unknown form of nonstationarity. Let's define function $J^*$ as the optimal performance of a learning agent and $J^{**}$ as a suboptimal performance that started after the distribution shift. It is important to be confident that after some time steps, the performance will adjust, i.e.:
\begin{equation}
\label{eq:eq6}
 \lim_{t\to \infty} \big( J^*_{i,t}-J^{**}_{i,t} \big) \leq \sigma
\end{equation}
where $\sigma$ is some threshold that defines an acceptable performance value. Let's define the loss function $L_{M_z,t}$, for the episode $M_z$  as the difference between optimal performance $J^*_{i,t}$ and suboptimal performance $J^{**}_{i,t}$, $L_{M_i,t}= J^*_{i,t}-J^{**}_{i,t} $  and rewrite ( \ref{eq:eq6}) as follows:

\begin{equation}
\label{eq:eq7}
\lim_{t\to\infty}L_{M_i,t} = L_{M_i,t}^* \leq \sigma
\end{equation}
 For safe adaptation, two conditions are important: the performance must get back to normal quickly, and besides, during the adjustment, safety constraints must be satisfied. For the first condition, we can find the time of adaptation, i.e time that it takes an agent to get to the normal performance, if we use \ref{eq:eq7}. We can define some adaptation function $f_{adap}(\pi, t)$ that finds the policy $\pi$ that would return the minimal time  of adjustment $t_{min}$, i.e., such a time $t$ at which the loss $ L_{M_i}$   would satisfy the following conditions:  
 \begin{equation}
 \label{eq:eq8}
 f_{adap}(\pi, t) = L_{M_i,t} -\sigma\leq o
\end{equation}
 The naive approach suggests that the adjustment to nonstationarity occurs when the speed of change of $J^*$ is equal to the speed of change of $J^{**}$. So, we can formulate an adjustment as specified in equation \ref{eq:eq9}.
 
 \begin{equation}
 \Delta J_{i,t}^*= \Delta J_{i,t}^{**}
 \label{eq:eq9}
\end{equation}
 Then we can formulate the adjustment time $t_{adj}$ as a minimal time that satisfies equation 9.
 \subsection{Safety consideration}
 Safety specification for continual RL in nonstationary environments has several aspects that are different from traditional, stationary safe RL specifications. One difference is that after the deployment, the system should remain safe for the lifetime of the RL learning agent, including the time required for adjustment to the distribution shift. The second difference is that constraints for the future episodes are often unknown at the time of making the constraints specification. Usually, we can only set up constraints for the initial episodes. There is no guarantee that these initial constraints will remain valid for future episodes. For example, if we set the speed limit to 40 miles per hour for an autonomous car based on testing environment conditions, this limit can be dangerous on the highways where the speed limit is higher, or during off-road driving, where the speed should be much lower to remain safe. Even in a familiar environment speed limit can be affected by traffic or road conditions such as rain or snow.\\
 In nonstationary environments, constraints should take the environment, or so-called context, into account. That is true especially for the hard, instantaneous per-state constraints.
 \subsubsection{Formulation safety constraints}
 
  One way of formulating the constraints includes formulating a constraint function $f_{_C}$ and setting some threshold value $\sigma$. 
\begin{equation}
\label{eq:eq10}
 f_{_C}(s,a) \leq\sigma
\end{equation} 
  States, state-action or state-cost pairs are typically used as arguments \cite{berkenkamp2023bayesian}. 
However, in traditional stationary MDPs, states or state-action, state-cost pairs do not reflect information about the dynamics of the environment caused by a distribution shift. As the environment evolves, this can cause degradation of the safety constraint function unless it updates with the environment. The same is correct regarding the threshold. 
One solution can be exact knowledge of the constraint function dynamic and the dynamic of the threshold change. In a simpler case, we can design a deterministic function $f_{cd}(t)$ and a threshold $\sigma(t)$  such that $f_{cd}(t) \leq\sigma(t)$. In a general case, such models are imprecise and difficult to build, as for continual learners at design time, many constraint requirements are usually uncertain and difficult to predict. Another solution offers to use knowledge of the environment context to build constraints \cite{berkenkamp2019safe}. Utilizing the context allows us to build adjustable data-driven constraints that can change with the environment.
We can formulate context-based constraints as follows
\begin{equation}
 f_{cvar}(s_z,a_z,r_z)   \leq\sigma'
\label{eq:eq11}
\end{equation} 
where $z$ is a context variable that reflects the environment change, i.e., change in the distribution of the parameters 
in episodic MDPs. 
\subsubsection{Safety during the adjustment to DS}
Another way to build data-driven constraints - to use statistical parameters of the environment. During the adjustment to the distribution shift, we don't have access to the context data, but we can use, for example, variational budgets, i.e. maximum number of violations, of the performance thresholds  \cite{chandak2020lifelong,wachi2020safe, wachi2024survey}.  
\par We can also use the adjust time and adjustment speed (see section 3.2) to set up the constraints during the adjustment period.
\subsection{General problem formulation}
 We formulate the problem of safe continual online learning as finding a policy $\pi$ that maximizes cumulative reward during the lifetime of the learning agent 
 \begin{equation} 
 \label{eq12}
  \max_{\pi\in \Pi} J_{\pi} =  \max_{\pi\in \Pi}\mathbb{E}_{\pi} \left (\sum_{i=1}^\infty \sum_{t=1}^\tau \gamma^t *r_{t} \right )        
\end{equation} 

while satisfying: 
 
\begin{equation}
\label{eq13}
    \begin{cases} general\hspace{0.2cm} safety \hspace{0.2cm}constraints:\begin{cases} f'_{c} \leq\sigma' \\ 
          f_{cvar}(s_z,a_z,r_z)\leq\sigma_{var}\\
           \end{cases}\\
           \\
      adjustments \hspace{0.2cm} constraints: t_{adj} \leq {\sigma_{adj}}
    \end{cases}\
\end{equation}
where $f'_{c}$ and $f_{cvar}$ represent constant and variable constraint functions and $\sigma'$ and $\sigma_{var}$ constant and variable thresholds correspondingly, $z$ is a context variable, $t_{adj}$ and $\sigma_{adj}$ represent time of adjusment and adjustment threshold value. Adjustment constraints show the upper bound for adjustment time. We believe that adjustment is an important part of safety and that adjustment constraints should be considered together with safety constraints for COSRL methods. We will provide more details on constraints in \ref{sec: Taxonomy of constraints}.

\section{Challenges of continual safe online reinforcement learning} 
  In this section, we introduce just the COSRL problems that are relevant to our study. A detailed description of all challenges is far beyond the scope of this review. 
 Some of the most significant challenges are:\\
 
1) Small sample size is known as one of the biggest problems of online RL. Online RL deals with a consecutive sampling that represents a sample efficiency challenge, as the algorithm can sample only one trajectory $h_i$ from each episode $i$. It is not physically possible to go back in time and have more trajectories for the same episode. \\

2) Another challenge is policy optimization, i.e., selecting the best policy, the policy that maximizes the performance. Even if policy performance trends are generated, we have to compare them to select the best one. To be able to do that, we need some unbiased estimator mechanism that does not depend on nonstationarity. Besides, because of the interplay between reward and the cost function convexity, the assumption holds only for some specially designed laboratory settings.
That makes it difficult to apply traditional gradient-based methods for optimization, which require the convexity of the objective function. \\

3) Setting the safety constraints is not a trivial task in online RL that acts in a nonstationary environment. In most cases, safety constraints are unknown or partially unknown at the deployment stage. Moreover, constraints can be time-dependent or data-driven and depend on the distribution of the data.\\ 

4) Scalability is one of the most serious challenges for the COSRL practical application. Many of the theoretical approaches are resource-hungry and work well only in small-scale applications. All those challenges ask for specific solutions that are, in most cases, different from algorithms based on the stationarity assumption.\\

\section{Current Solutions}

Online adaptive algorithms usually have the following structure. A decision model maps the input variables to the target, and the learning algorithm specifies how to build a model from a set of data instances. Online adaptation procedure usually includes three steps: prediction, diagnosis, and update.  When new input data $x_t$ arrives, a decision model $y$ makes a prediction: $y(x_t) \rightarrow y^p(x_t)$. At the diagnosis stage, we can sample the results, the true labels $y^T(x_t)$,  compare them with the prediction, and calculate the loss, that is, the difference between the prediction $y^p(x_t)$ and the true label $y^T(x_t)$. At the final stage, we can use $(x_t,y^p(x_t))$ and the loss to update the decision model.\\
We are going to consider current solutions in COSRL based on the adaptation techniques of the algorithms. Adaptation speed indicates how quickly the RL agent can recover performance and safety after facing the distribution shift.\\

Most of the considered methods ensure safety by modifying the existing, based on the stationarity assumption, safe learning methods to adjust them to a nonstationary setting. Safety level guaranteed by the algorithm is closely related to the adaptation mechanism. Therefore, we provide the following taxonomy of safe online reinforcement learning methods based on the adaptation model: passive adaptation, reactive adaptation, quick proactive adaptation, and proactive adaptation.
\subsection{Passive safety adaptation}

\textbf{Passive safety adaptation} proposes that to stay safe during and after the distribution shift, the agent should stay in the region with the familiar data distribution or within the safe state set, which is not going to change despite the nonstationarity.
The assumption is that the distribution shift doesn't completely cover the data set, and there are some regions of the data with the familiar distribution. The only thing that the algorithm should do is either to stay within the region with the familiar distribution \cite{daneshmitigating,yang2019safety, ohnishi2019barrier,berkenkamp2017safe} or to stay within some safe set region that can be determined during the offline training \cite{berkenkamp2017safe,prajna2004safety}.  
Safe RL methods for regions for the familiar stationary data distributions on which the agent was already trained are well known, and we refer our readers for more information to \cite{gu2024review,kim2020safe, brunke2022safe}, and the comprehensive work of Garcia and  Fernandez \cite{garcia2015comprehensive}.


This approach eliminates the necessity for performance and safety adaptation because it keeps the agent in the distribution. The most important part of the algorithm is the ability to recognize distribution shift and stay away from unfamiliar data.  Operating in the familiar region can narrow the applicability of the algorithm, but guarantees the minimal performance and hard safety constraints satisfaction in case the distribution shift(DS) does not affect all the data. However, if DS changes the whole data, the proposed algorithm can not be applied. Even a slow but gradual change can affect all the data over time. The open question is whether the new data provide a chance for better performance if the performance and the constraints were adjusted to the DS. Can the adjusted algorithm work better if it stays in the familiar data distribution?

\subsection{Reactive safety adaptation}
\textbf{Reactive safety adaptation} approaches propose conservative online adaptation of the policy performance while satisfying safety constraints. Technically speaking, every online on-policy RL algorithm can adjust to the new data as soon as it updates the model.  Those methods make a forecast of the policy improvement, make the sampling, evaluate the loss, and update the value function. Despite making the prediction of local policy improvements,  based on the current state and action, such adaptation is passive and assumes slow change of the environment, and has a lag in performance. These algorithms do not estimate the upcoming changes in the context and the dynamics of the context changes. One of the biggest drawbacks of on-policy algorithms is that they are generally less sample-efficient than off-policy algorithms. They work well if data is available. However, during online learning, the data is scarce. 
Chandak et al.\cite{chandak2020towards} proposed one of the first model-free online safe policy improvement RL algorithms, based on sequential time series analysis. The algorithm considers nonstationarity in the reward function $R(x)$ and in the transition probability function $P(x)$. Adjustment of the performance of the algorithm is based on incremental and safe policy improvement. The algorithm generates predictions of the performance of new policies and then selects the best candidate policy based on sequential testing of the performance of the forecasted candidates. To generate new policies and forecast their performance, the algorithm uses counterfactual analysis of the past performance, searching for the answer to the question of what could have been the performance of the proposed policy during the previous episodes if this policy had been applied. For selecting the best candidate, the algorithm consecutively evaluates the performances of the candidates, i.e, it consecutively evaluates the hypotheses of how the performance of the algorithm was improved if one or another had been implemented in the past.  To select the best candidate, the algorithm evaluates the uncertainty related to the performance of each generated policy based on the performance variance of the policy. Each generated policy has a different performance variance. While some policies have better performance than others, they also can have higher variation. The conservative approach proposed by the authors suggests that the best candidate is the policy that has the highest low variance of performance. If the candidate has better performance and lower variance of the performance than the current policy or the so-called safe policy, then the current policy will be updated by the policy candidate. There are several important points to highlight about the algorithm. The first one is that authors \textbf{define safety as improved performance}. The policy is safe if it increases the performance of the algorithm. While such an approach is very comfortable for optimization, as we need to optimize just one function - the performance, it does not take into account any constraints that are not connected with the performance.  The second one is the assumption of the smoothness of the policy performance function. That means that policy performance can't suddenly change and should satisfy Lipshitz's continuity assumption. That means that in the considered setting, the policy performance would change slowly from episode to episode. Slow adaptation can compromise the safety if nonstationary changes occur quickly. The third important point to note is that authors use linear regression as the prediction mechanism of the future performance and assume the linear nature of policy improvement.   Both assumptions are strong, and while being comfortable, simplification to some extent limits the applicability of the algorithm, as real-life non-stationarity can change very fast.
\par Ding et. Al \cite{ding2023provably} investigated a more practical setting where transition, reward, and constraint functions can vary over time. The proposed online on-policy algorithm considers nonstationary episodic MDPs. The variations can happen in an episode or between episodes. They formulated the problem as a constrained optimization problem and proposed the primal-dual approach for optimization. As soon as the variation of the constraint function is unknown but can be rather large, to provide safe exploration, the authors assumed either knowledge of episodic variational budget or knowledge of local variational budget, i.e, knowledge of the constraints for a smaller part of the episode named epochs, which are back-to-back time intervals that can span several episodes.  The proposed periodically restated optimistic policy evaluation algorithm, which is based on the online mirror descent optimization method, is similar to well-known optimization methods that are widely used for stationary CMDP, such as  TRPO and  PPO optimization methods. Notably, the proposed optimization uses KL divergence regularization that takes into account the distance between active and proposed policies.   Periodic restart is a mechanism designed to update the model in order to adjust it to non-stationarity. The proposed method introduces expected cumulative constraints to nonstationary CMDPs. The work established dynamic regret bound and constraint violation bound for the algorithm in linear kernel and tabular kernel settings. As in the previous work \cite{chandak2020towards}, the proposed constraints are soft constraints that hold in expectation.
\par Wei et al. \cite{wei2023provably} considered similar settings of non-stationarity in the episodic CMDP framework as in  \cite{ding2023provably},  but proposed a model-free algorithm and a modified version of safety constraints that do not require prior knowledge of budget constraints.  It implies knowledge of the constraint function - the budget, but the threshold of the constraint function can be calculated automatically. The authors proposed double restart as a model adaptation measure. 
 A different adaptation technique was proposed by Car et al. \cite{carr2023safe}, who extended the applicability of shields in deep reinforcement learning. The proposed algorithm can guarantee safety during training in a shielded environment and during testing with the shield off. This approach utilizes a POMDP model to represent nonstationarity. The algorithm builds a belief support as a state estimator and then uses a state estimator and a reach-avoid specification, given a priori, for safe action selection. Belief support is formed based on the history of actions and observations. So, the updating of the belief support, which is a foundation of the shield, is reactive rather than proactive, and that can slow down the adaptation.
\subsection{Quick safety adaptation}
Quick safety adaptation includes methods that actively try to figure out the context model, i.e, find the set of distributions of the environment, and based on the context, predict the best performance policy and best safety policy. Nonstationary MDP (NSMDP) is represented in an episodic setting \cite{chen2021context,khattar2024cmdp,wen2022improved}  where the transition function and reward can vary. The main idea is to find context variables $z_i$, or so-called latent variables, that can represent the distribution 
at each particular episode $i$. Then, regular CMDP can be modified as follows: $CMDP(S, R, P, A, C, Z)$. In the general case, latent variables can parametrize all the parameters of the CMDP process, and a particular hidden variable is responsible for each parameter: $CMDP(S_{Z_s},R_{Z_r},P_{z_p}, A_{Z_a}, C_{Z_c})$. Knowledge of latent variables can provide a tailor-made solution for finding a policy that fits best for each NS-MDP episode. 

Chen et al.\cite{chen2021context} proposed a model-based context-aware safe algorithm that incorporates domain knowledge to guarantee safety. This method uses a probabilistic latent variable model for figuring out the context, based on amortized variational inference analysis.  The context is then used to improve the prediction of the next state. Authors use a sample prioritizing method to improve the sample efficiency of rare unsafe events during training. Safety is implemented with risk-averse decision-making with constrained model predictive control.  Remarkably, alongside the traditional formulation of CMDP, the authors use two sets of probabilistic state constraints to modify optimization criteria. The first set of constraints concerns the states predicted by the context-parametrized function $f_z$, such as $P(f_z(s_i)=s_{i+1}\in S_{unsafe}) < \sigma$, where P is a probability.  The second set of state constraints is predicted by the traditional function $f$  not parametrized by context $P(f(s_i)=s_{i+1}\in S_{unsafe}) < \sigma$. Such double constraints were set for the case if the context variable method fails.  Solving multiple constrained tasks that include latent variables is intractable. To overcome this challenge, the authors propose sample-based model predictive control. They propose a modified objective function that includes both state constraints and action constraints.  Action constraints are then evaluated using the conditional value at risk (CVaR) approach. To guarantee risk-averse behavior, only actions with high value at risk are preferred. The algorithm assumed knowledge of safe states. Despite theoretical discussion of two sets of probabilistic per-state safety constraints, this algorithm does not offer hard safety constraint satisfaction because solving the optimization problem in the proposed constrained setting, parametrized by context variables, is hard.  At the same time, the considered adaptation method based on context learning shows the way for quick adaptation. Still, the question remains how to formulate hard safety constraints in a non-stationary MDP to make the solution trackable.\\
One of the most promising directions for quick, safe adaptation is online safe meta-learning adjusted to the CMDP framework. Reinforcement meta-learning algorithms were designed to accelerate learning and to provide quick adaptation to unfamiliar tasks \cite {finn2017model,hazan2016introduction,xu2024meta, acar2021memory, finn2019online}.
However, the introduction of constraints and the adjustment of meta-learning to the online CMDP framework is not a straightforward task.  The nonconvexity and interdependence of rewards and constraints make it challenging to apply existing optimization approaches \cite{hazan2016introduction}. Besides, it is impossible to apply many of the existing online reinforcement learning algorithms that assume an unbiased estimator of the loss function \cite{khodak2019adaptive}, as finding of optimal global policy for online  CMDP is not realistic. 
\\
Khatar et al. \cite{khattar2024cmdp} proposed a meta-safe reinforcement learning framework (Meta-SRL) that addresses these fundamental problems. Meta-SRL extends the meta-learning concept to safe reinforcement learning and provides a low-regret online learning framework. The proposed algorithm has two layers - a meta-algorithm as an external layer and a within-task algorithm as an internal layer. The purpose of the within-task algorithm is to learn each particular task, and the purpose of the meta-algorithm is to tune up the work of the within-task algorithm.  Within-task algorithm is responsible for the adaptations to each task, including adjustment of the learning rate and policy improvement, while the meta-learning part is responsible for making rate adjustment and policy improvement more quick and precise based on task similarities. The method uses CRPO Xu et al.\cite{xu2021crpo} for internal layer optimization.  Safety constraints are presented as cumulative expectation constraints. That type of constraint setting is very similar to \cite{wei2023provably, ding2023provably} 
variational budget constraints. The method makes Lipschitz assumptions about the value function and policies.

Lu Wen et al.\cite{liu2021policy} proposed a meta-RL framework that introduces safety during the pre-adaptation stage and the performance guarantees during the adaptation stage of the RL agent. The work proposed a safe off-policy meta RL method, PEARL+, that adds safety consideration to meta reinforcement learning via the probabilistic context variables, and is built on the top of well known PEARL method, proposed by K. Rakelly et al. \cite{rakelly2019efficient}. \cite{liu2021policy} offers prior regularization terms in the reward function and adds Q network for recovering the state-action value under prior context assumptions. The primary objective of the method is to optimize the policy for both prior (pre-adaptation) safety and posterior (after-adaptation) performance. Policies are assumed to be safe if they do not bring the agent to the unsafe state. The work assumed knowledge of a safe state set.\\

\subsection{Proactive safety adaptation}
Considered context-based methods guarantee fast adaptation but make no predictions about the dynamics of the contexts. Identification of the context change goes reactively. That can result in sample inefficiency, slack of performance, imprecise predictions, and safety violations. Knowledge of the context dynamics could help to alleviate these problems by predicting non-stationarity.  If we have a time series of context variables$z_1, z_2,..., z_i$  we can predict the posterior $z_{i+1}$ by time series analysis.
Zhenyuan Yuan1 et al. \cite{yuan2025all} propose to solve online learning of control policy parameters by modeling temporal relations between tasks as a Markov process and solving the online policy optimization problem considering tasks as states and control policy parameters as actions. A high sample efficiency algorithm, FTLPP, was developed based on that concept.  In contrast with Khatar et al. \cite{khattar2024cmdp}, this method does not require calculation of the gradient of the objective function. Calculation of the gradient requires the generation of the trajectories at each time step, which is resource-intensive. The proposed method solves this problem. In the safety part \cite{yuan2025all}, introduce a policy-masking approach that is different from the safety mechanisms previously considered in our survey. The key idea of policy masking is building a so-called masking function for the control policy that, with high probability, will guard the control policy from making unsafe steps within the masked control policy space. Building a masking function allows to transform constrained policy optimization to unconstrained over the masked policy and sufficiently simplifies the optimization task. The idea of building a masking function is based on the unsafe state set known a priori for each task. If the unsafe set is known, we can calculate the one-step backward set. All actions that lead to $\epsilon-step$ backward set and all the states in this set are considered to be unsafe. Then, for each policy $p$ for the task $i$, we can find such a function  $\mu$ that will redirect the policy to safe action and safe state sets.  This safety approach can guarantee cumulative constraint satisfaction for each task. Technically, the idea of masking unsafe policies can be realized by implementing a retrainable safety layer in a deep neural network. A similar idea, but for a  stationary MDP, was proposed by Dalal et. al \cite{dalal2018safe}. The biggest challenge with that type of safety setting is that usually, the safe state set for the task is unknown. For online learning, we can expect some domain knowledge for the first tasks, but it is unrealistic to assume knowledge of the safe states for all the future tasks, especially in an infinite MDP. It would be helpful to have a mechanism able to predict safe states for unseen tasks based on the context or on task similarities.

\section{Details of the algorithms}
 The algorithms we discussed provide a wide range of approaches to ensure online learning safety. This section summarizes proposed solutions across the following five areas:  learning, adaptation, optimization, safety consideration, and the type of environment (see Table \ref{tab:my_label1} and Table \ref{tab:my_label2}).  We also provide details on the theoretical foundation of the algorithms, including the main assumptions of the algorithms (Table \ref{tab:my_label3}).

  \begin{table} [t!]
    \centering
    \caption{Safe RL under Non-stationarity Methods. Part 1}
     \begin{scriptsize}
        \label{tab:my_label1}
      \begin{tabular}{|p{7em}|p{4em}|p{3em}|p{4em}|p{4em}|p{4em}|p{4em}|p{4em}|p{4em}|p{3em}|p{4em}|p{3em}|p{3em}|p{3em}|}\hline   
          \multicolumn{2}{|c|}{}&  \multicolumn{4}{|c|}{Learning}&      Ada-ptation&\multicolumn{2}{|c|}{Optimization}&\multicolumn{3}{|c|}{Safety Consideration}&\multicolumn{2}{|c|}{Environment}\\\hline
 Citation&     Algo-rithm& MDP type& Type of Learner&  Type of Policy Update&  Explo-ration Strategy&      Ada- ptation Mecha-nism&Optimi-zation Method&Optimi-zation Criteria&Risk criteria&Type of Safety Constraints&   Type of Guarantees&Space complexity& NS Type\\\hline\hline
 \multicolumn{14}{|c|}{Passive safety adaptation}\\\hline
Motoya\cite{ohnishi2019barrier} & Safe Learning Framework  &non MDP &model-based & on-policy& greedy&recovery of safety in
the sense of Lyapunov stability after violations &sparse optimization techniques & max perfor-mance while satisfying constraints&safety barier violation &safety barier certificate &in expectation  &CS&R, P\\\hline
Berkenkamp\cite{berkenkamp2017safe} &  SAFE LYAPUNOV LEARNING &non MDP type &model-based & on-policy&greedy & recovery of safety in
the sense of Lyapunov stability after violations&prima-dual methods for solving Lagrangian &max performance while satisfying constraints &safety barier violation &Sfety barier certificate & proba-bilistic&mostly
for DS, but can be ad-justed for CS&R,P,A\\\hline\hline
\multicolumn{14}{|c|}{Reactive safety adaptation}\\\hline
Chandak\cite{chandak2020towards} & Seldo- nian  & epi-sodic, consequtive MDP, (NS-MDP) &model-free, PB& off-policy & Highest low variance of return policy&  relies only on estimates of future perfor-mance,
with associated confidence intervals&a gradient-based & max lower bound of the policy perfor-mance&perfor-mance level &Cons-traints on policy perfoprmance &proba-bilistic in expectation &CS&Exo-genous. R, P \\\hline
Ding\cite{ding2023provably} &PROPD-PPO   &epi-sodic, conse-qutive MDP, (NS-MDP) &model-based &on-policy & safe exploration&a periodic restart & prima-dual, gradient&Max performance while satisfying variational budget & Var. budget violations& Non-sta-tionary constrain function that varies over the episodes&in expectation&DS, TDS &R,C,P\\\hline
Wei\cite{wei2023provably} & Algo-rithms 1,2,3,4  & epi-sodic NS-CMDP& model-free, PB, simu-lation-free & on-policy& optimism in the face of uncertainty& periodic restart, double restart &primal-dual mirror descent & Max performance while satisfying safety constraints& Var. budget violations& Non-sta-tionary const-raint function that varies over the episodes& in expectation &mostly for DS, TDS, but can be adjusted to CS &R,C,P
\\\hline
Qiu \cite{qiu2020upper} & UCPD  &epi-sodic NS-CMDP &model-free, PB & on-policy& optimism in the face of uncertainty& on-policy update& primal-dual & maximize performance while satisfying  cost constraints& varia-tional budget violation&non-statio-nary cost function & cumu-lative& DS&R,C\\\hline
Car\cite{carr2023safe} & flexible  &POM-DP&flexible &on-policy & greedy&shielding & flexible framework, incl DQN, DDQN, REINFORCE e.t.c&maximize performance while satisfying reach-avoid specification& viola-tion of the reach-avoid specification&reach-avoid specifications & hard constraints (probabilistic per state con-stra-ints)& DS, CS &P and R beliefs\\\hline
      \end{tabular}
      \end {scriptsize}
     \label{tab:placeholder}
 \end{table}

\begin{table} [t!]
\centering
    \caption{Safe RL under Non-stationarity Methods. Part 2}
     \begin{scriptsize}
        \label{tab:my_label2}
      \begin{tabular}{|p{7em}|p{4em}|p{4em}|p{4em}|p{4em}|p{4em}|p{4em}|p{4em}|p{4em}|p{3em}|p{4em}|p{3em}|p{3em}|p{3em}|}\hline   
          \multicolumn{2}{|c|}{}&  \multicolumn{4}{|c|}{Learning}&      Ada-ptation&\multicolumn{2}{|c|}{Optimization}&\multicolumn{3}{|c|}{Safety Consideration}&\multicolumn{2}{|c|}{Environment}\\\hline
 Citation&     Algo-rithm& MDP type& Type of Learner&  Type of Policy Update&  Explo-ration Strategy&      Ada- ptation Mecha-nism&Optimi-zation Method&Optimi-zation Criteria&Risk criteria&Type of Safety Constraints&   Type of Guarantees&Space complexity& NS Type\\\hline\hline

 \multicolumn{14}{|c|}{Quick safety adaptation}\\\hline
Chen\cite{chen2021context} &  CASRL & NS-CMDP&model-based & on-policy& Gaussian type, risk-averse decision making& meta-learning: context-aware probabilistic latent variable model&prima-dual, predictive control & Max performance while satisfying safety constraints& Condi-tional value at risk (CVaR)&risk averse decisions& Proba-bilistic constraints & DS, CS &P\\\hline
Khatar\cite{khattar2024cmdp} &Meta-SRL   & NS-CMDP (Inexact CMDP-within-online framework)& model-free, PB& on-policy&flexible &meta-learning   &online gradient decent (OGD) & Max performance while satisfying  cost constraints& viola-ting the threshold&cost function &expe-cted constraints (expected constraint violation) & DS, TDS, CS &Diffe-rence between tasks  : P, R, S, A,  policy similarities\\\hline
Wen\cite{liu2021policy} &PEARL PLUS& NS-MDP  &flexible &off-policy & tempo-rally extended exploration at unseen tasks &meta-learmimg: Proba-bilistic embeddings for actor-critic RL &Soft Actor Critic (SAC)  &maximize performance while satisfying  cost constraints& viola-tion of the threshold& cost function& proba-bilistic &CS, DS, TDS &task distribution shift\\\hline
 \multicolumn{14}{|c|}{Proactive safety adaptation}\\\hline
Yuan1\cite{yuan2025all} &  masked FTLPP framework &NS-MDP & model-based task transition&off-policy& safe exploration by policy masking  &meta-learning  & gradient-free or gradient-based&maximize performance while satisfying  reach-avoid specification& viola-tion of the reach-avoid specification &cost function, reach-avoid specifications &hard constraints (probabilistic per state con-stra-ints) &mostly for DS, TDS  but can be adjusted for CS &R, initial state distribution, time horizont T\\\hline\hline
     \end{tabular}
     \end {scriptsize}
     \label{tab:placeholder}
     
 \end{table}

\subsection{Learning}
This subsection summarises details of the RL algorithm and includes four subgroups: a) MDP type, b) the type of learning, b) the type of policy update, and c) the type of exploration. \\
\textbf{MDP type} specifies MDP model and includes the following abbreviations: NS - nonstationary, ep - episodic, seq - sequential.
\textbf{The type of learning} specifies whether the algorithm is model-based (model-based) or model-free (model-free). Model-based algorithms try to learn the MDP model first, i.e., they learn the mapping function $P(x)$ that defines the transition probability between states, while model-free algorithms do not make any assumptions about the model of the environment.  Model-free algorithms can be value-based (VB) or policy-based (PB). While value-based algorithms define state value, i.e., how good it is to be in a particular state, policy-based methods evaluate directly the policy, i.e., the strategy that maximizes the value of the expected cumulative return. Softmax learning (S-M) provides a combination of value-based and policy-based learning methods \cite{goodfellow20166}.\\
\textbf{Type of policy update} characterizes the way for policy improvement. On-policy algorithms (on-policy) learn the best policy by updating the current policy every time the agent receives feedback, while off-policy (off-policy) algorithms learn the optimal policy by observing the behaviour of the current policy of the algorithm.  Knowledge of policy update type is important because on-policy algorithms in general require more samples to learn. On the other hand, with a sufficient number of samples, the algorithm has a good chance to adjust to a new distribution.\\
\textbf{The type of exploration} section includes exploration strategy and exploration methods when they are available. As it is not always easy to separate methods from strategy and to provide an exact formal classification and description of both, given the space provided by the table, we will provide a brief verbal description of exploration rather than a formal classification. We believe this format would be easier to follow for potential readers.

\subsection{Adaptation mechanism} 
This section describes the key idea of adjustment to the distribution shift mechanism.
\subsection{Optimization} \
 This section considers optimization details and includes two subsections: optimization methods and optimization criteria.
 \textbf{Optimization Method} section provides a brief verbal description of the optimization method applied. 
 \textbf{Optimization Criteria} subsection provides a brief verbal description of the optimization goals.

\subsection{Safety Consideration} includes three subsections: risk criteria, type of safety constraints, and type of safety guarantees. 
\textbf{Risk criteria} subsection explains the safety idea proposed by the method. Again, due to the variety of safety approaches, we provide a verbal description rather than a formal definition, as providing a comprehensive formal definition of safety and risk criteria is far beyond the scope of this review. On the other hand, we believe that verbal description makes it easier to follow for potential readers.  
\textbf{Type of Safety Constraints} describes the mechanism of safety constraints implemented by methods. Safety constraints, in the methods under review, are set by four different ways, including: a) setting safety barriers based on the Lyapunov stability concept, b) setting improvement criteria on policy update, c) setting a separate constraints function, or so-called cost function, and, finally, d) setting the constraints based on reach-avoid specification.  Constraints mentioned in $a$ and $d$  are both based on knowledge of a safe set a priori, but the mechanism of constraints is quite different; that's why we specify them as two different types. 
\textbf{Type of Safety Guarantees} describes types of safety guarantees provided by methods. Briefly speaking, all safety guarantees constraints in the methods under review can be divided into three groups: expected, cumulative, and probabilistic, including probabilistic per-state or, so-called, hard guarantees. For more details on those types of constraints, we refer our readers to \cite{wachi2024survey}.
.
\subsection{Environment and non-stationarity type}
Includes two subsections: a space-complexity parameter that characterises the state space and a nonstationarity description.\\ 
\textbf{State complexity} shows to what types of environments the algorithm was designed and serves as a good indication of the potential applicability of the algorithm. Space complexity includes the following parameters: a continuous state space (CS), a discrete state space (DS), a tabular discrete state space (TDS), and an infinite discrete state space (IDS). 
\textbf{Nonstationarity Type} describes the following types of nonstationarities: nonstationarity in the transition probability function - (P), nonstationarity in the reward - (R), nonstationarity in the actions - (A), nonstationarity in constraints - (C), or nonstationarity between task similarities, including combinations of P, R, A, C  \\
Table \ref{tab:my_label3} represents the key assumptions of the methods.
 \begin{table}
    \centering
    \caption{Assumptions of the methods}
     \begin{scriptsize}
      \label{tab:my_label3}
      \begin{tabular}{|p{7em}|p{10em}|p{25em}|}\hline   
           \multicolumn{1}{|c|}{Autors}&  
           \multicolumn{1}{|c|}{Algorithm}&      
           \multicolumn{1}{|c|} {Asumptions}\\\hline
 \multicolumn{3}{|c|}{Passive safety adaptation}\\\hline
Motoya\cite{ohnishi2019barrier} & Safe Learning Framework  &Lipschitz continuity of the control barrier function  \\\hline
Berkenkamp\cite{berkenkamp2017safe} &  SAFE LYAPUNOV LEARNING & Lipshitz continuity of the objective and utility functions\\\hline\hline
\multicolumn{3}{|c|}{Reactive safety adaptation}\\\hline
Chandak\cite{chandak2020towards} & Seldonian &Lipschitz smooth performance of the objective function    \\\hline
Ding\cite{ding2023provably} &PROPD-PPO   &Local variational budget is known, or strict feasibility thresholds are known 
\\\hline
Qiu \cite{qiu2020upper} & UCPD  & Existance of the solution for Lagrangian dual function\\\hline

\multicolumn{3}{|c|}{Quick safety adaptation}\\\hline
Chen\cite{chen2021context} &  CASRL & a. The environment is episodically consistent, and changes happen at the beginning of each episode b. Safe state set $S_{safe}$ and safe action set $A_{safe}$ are known a priori \\\hline
Khatar\cite{khattar2024cmdp} &Meta-SRL & Lipshitz continuity of policy parameters
   \\\hline
Wen\cite{liu2021policy} &PEARL PLUS&  \\\hline
Car\cite{carr2023safe} &  & Knowledge of reach-avoid specifivation  \\\hline\hline
\multicolumn{3}{|c|}{Proactive safety adaptation}\\\hline
Yuan1\cite{yuan2025all}  &masked
FTLPP
frame-
work&a. Latent Markov Process for task transitions b. unsafe states set $S_{unsafe}$ is available a priori\\\hline

     \end{tabular}
     \end {scriptsize}
     \label{tab:placeholder}
     
 \end{table}

\section{Taxonomy of constraints}
\label{sec: Taxonomy of constraints}
Based on review data, we categorise constraints as follows. One approach formulates safety as an improved performance of the algorithm and puts constraints on the performance of the algorithm in the following way: each new policy update should improve the performance of the algorithm. Another approach explicitly takes safety into account and considers not only constraints on performance improvement but also some additional constraints. Based on a state-of-the-art survey, safety constraints can be divided into three groups:
 \begin{table}
    \centering
    \caption{Taxonomy of constraints}
      \label{tab:my_label99}
 $ \textit{Constraints} \hspace{0.2cm} 
 {\begin{cases}   
 Performance-based\\ 
 \hspace{3.3cm}\\
 Safety-based \\ 
\end{cases}
 \ } 
 {\begin{cases} Predefined\\ \\ 
 
 Partially\hspace{0.3 cm} adjustable   
 \begin{cases}  
 Data-driven\\ 
 Time-dependent 
 \end{cases}\\ \\     
 Adjustable  
 \begin{cases}  Data-driven\\ 
 Time-dependent 
 \end{cases} 
 \end{cases}}$

     \label{tab:placeholder}    
 \end{table}
predefined constraints, partially adjustable, and adjustable constraints.  Predefined constraints represent domain knowledge and are assumed to be known from the beginning. These include knowledge of the safety function and knowledge of the threshold. Both of them are given and don't change during the training or after the deployment, and are the same for all episodes. For example, domain knowledge can be represented by knowledge of the unsafe states \cite{yuan2025all}, \cite{khattar2024cmdp}  
or by knowledge of safety function and knowledge of safety threshold \cite{chen2021context}.     
We classify the constraints as partially adjustable if either the safety function $f_{safe}(x)$ or the threshold $\sigma$ is adjustable but not both. Adjustable constraints are the constraints where both the safety function and the threshold are data or time-dependent. This is the most popular approach to setting constraints for safe online learning.  Adjustable and partially adjustable constraints can be of two types: data-driven and time-dependent. Data-driven constraints depend on some parameters of data distribution $P(s_i)$, e.g., the cost function can express cost variations defined as the cumulative number of violations of the threshold, and be upperbounded by some kind of indicator, such as a variational budget threshold known or unknown \cite{wei2023provably}. 
The value of the threshold can be given a priori or calculated based on some statistics over the episodes \cite{ding2023provably}, \cite{wei2023provably}.  
Time-varying constraints, e.g., safety function, that can change over time \cite{wei2023provably}. 
\section{Discussion}
In this paper, we reviewed algorithms designed for continual safe online reinforcement learning in non-stationary environments. Even though there is a big variation in setting constraints, optimization, and adaptation approaches, we can identify several common trends shared by all algorithms we analysed. The first is that a big part of the considered research is dedicated to quick adaptation to changes in distribution. Each new research contributes to the speed of adaptation, including proposed meta-learning algorithms and context-based models of adaptation \cite{chen2021context}, \cite{zhang2023adaptive}. The second finding is that 
despite the variety of proposed techniques, most of the works share the assumption of Lipshitz properties of the optimized function, i.e, continuity and bounded slope of the function. That is a strong assumption that would not hold in many practical settings. The Lipshitz property assumption limits the applicability of the methods to cases where nonstationarity changes slowly. Finding a way to get around this challenge will improve the prospects for quick adaptation. The third common trend is the types of constraints and safety guarantees. Except \cite{carr2023safe} and \cite{yuan2025all}, all considered methods provide only soft constraint guarantees. In other words, constraint violation is allowed within some limit, e.g., within variational budget \cite {ding2023provably}, \cite{wei2023provably}. One of the reasons for providing just probabilistic constraints \cite{chen2021context} or expected constraints \cite{wei2023provably, khattar2024cmdp} is the hardness of the optimization problem with variable constraints. The other reason is that setting statistic parameter-based constraints, i.e., variational budget \cite{ding2023provably, wei2023provably} or conditional value at risk \cite{chen2021context, ying2022towards}, is technically more straightforward to implement because of the availability of appropriate statistical parameters for each episode. Setting hard constraints requires knowledge of safe states and safe actions valid for the lifetime of the algorithm. However, in a nonstationary environment safe state and a safe action set can change over time. That means that in most cases, hard constraints are unknown. In the general case, safety functions and safety thresholds evolve and have time and data dependency.
Forecasting future safe states and safe actions is one of the potential directions of future research. 

Based on our survey, we believe that there are several \textbf{prospective avenues for research}. One promising direction is the development of \textit{context-based meta-learning algorithms, able to predict and proactively adjust to distribution shifts} based on the context dynamics. Developing such mechanisms would substantially cut the adjustment time. Another direction for future research is the \textit{development of adjustable data-driven hard constraints dependent on the context} of the environment. Finally, one more potential direction of research is the \textit{relation between the adjustment speed and safety constraints} for COSRL applications. Even though quick adaptation is a clear trend in the research we studied, we are not aware of any work that explicitly specified the speed of adjustment as a safety constraint. At the same time, adjustment speed is extremely important for safety-critical applications, and we believe that in the future, adjustment speed can be considered as a part of safety for COSRL methods.   

\section{Conclusion}
\label{conclusion}
In this work, we provided a review and  
taxonomy of the state-of-the-art COSR algorithms. We discussed some of the challenges for designing continual learning safe online algorithms and provided a categorization of the safety constraints that can work best for COSRL methods.
Based on our survey, we considered existing trends in the research and possible directions for future research in COSRL.  
Solving this problem would allow much wider acceptance of COSR in safety-critical applications.  

\bibliographystyle{plain}
\bibliography{bib}

\end{document}